\documentclass[runningheads]{llncs}
\usepackage{cite}
\usepackage{graphicx}
\usepackage{xcolor}
\usepackage{amsmath}

\usepackage[export]{adjustbox}
\usepackage{url}

\usepackage{booktabs}
\usepackage{wrapfig}

\begin{document}
%https://www.overleaf.com/project/5f7bdca5d90c8f00011c1cf7

\title{Detecting Anomalous Invoice Line Items in the Legal Case Lifecycle}
% \subtitle {(draft)}
\author{Valentino Constantinou\orcidID{0000-0002-5279-4143}  \newline
Mori Kabiri\inst{1}\orcidID{0000-0002-5745-0948}}
\authorrunning{V. Constantinou, M. Kabiri}
\institute{
InfiniGlobe, LLC, Newport Beach, CA 92660, USA\\ 
\email{Mori.Kabiri@infiniglobe.com}}

\maketitle              % typeset the header of the contribution
\begin{abstract}

The United States is the largest distributor of legal services in the world, representing a \$437 billion market \cite{USLegalServicesMarket}. Of this, corporate legal departments pay law firms \$80 billion for their services \cite{BenchmarkingReport}. Every month, legal departments receive and process invoices from these law firms and legal service providers. Legal invoice review is and has been a pain point for corporate legal department leaders. Complex and intricate, legal invoices often contain several hundred line-items that account for anything from tasks such as hands-on legal work to expenses such as copying, meals, and travel. The man-hours and scrutiny involved in the invoice review process can be overwhelming. Even with common safeguards in place, such as established billing guidelines, experienced invoice reviewers (typically highly paid in-house attorneys), and rule-based electronic billing tools (“e-billing”), many discrepancies go undetected. Using machine learning, our goal is to demonstrate the current flaws of, and to explore improvements to, the legal invoice review process for invoices submitted by law firms to their corporate clients. In this work, we detail our approach, applying several machine learning model architectures, for detecting anomalous invoice line-items based on their suitability in the legal case’s lifecycle (modeled using a set of case-level and invoice line-item-level features). We illustrate our approach, which works in the absence of labeled data, by utilizing a combination of subject matter expertise (“SME”) and synthetic data generation for model training. We characterize our method’s performance using a set of model architectures. We demonstrate how this process advances solving anomaly detection problems, specifically when the characteristics of the anomalies are well known, and offer lessons learned from applying our approach to real-world data.

\end{abstract}

\section{Introduction}

Most legal invoices are intricate, containing hundreds of line-items with detailed information about dates of service, quantity, rate, billing codes, and task descriptions. Generally, a line item could be a task or expense. The task or expense is identified by a standardized billing code that reflects the type of work done by the “timekeeper”, who is the law firm personnel who performed the task and reported the time.

\begin{figure}
\vspace*{-0.2in}
\includegraphics[width=\textwidth]{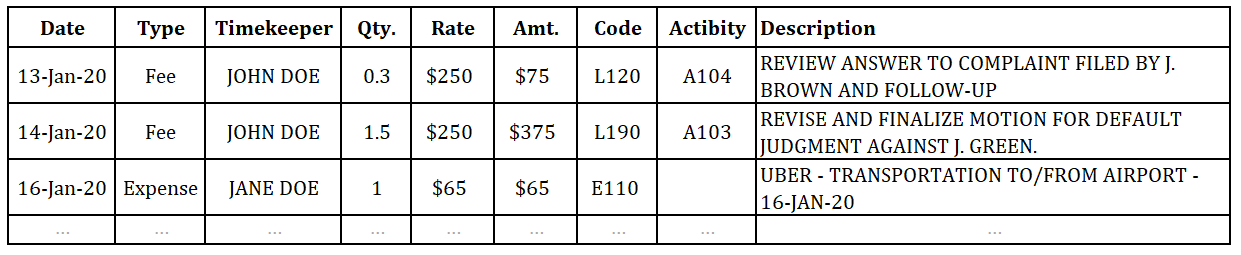}
\vspace*{-0.3in}
\caption{An example of line-items contained within legal invoices.} \label{category_codes}
\end{figure}

In the late 1990’s, \textit{e-billing} (applying a set of standardized codes to electronically-submitted invoices in an attempt to simplify and standardize the legal billing process and facilitate the ease of review and approval) was developed. Over the past two decades, most legal departments have shifted from paper invoices to e-billing tools to help validate invoices against their agreed terms with law firms and vendors, called “billing guidelines”. 

Based on an InfiniGlobe-led panel discussion with legal industry executives, most participants reported that of their law firms, ~80\% maintained “good compliance with billing guidelines” and ~20\% exhibited “poor billing hygiene”. Although the majority of law firms consistently submit accurate invoices based on actual work done, some submitted invoices contain irregularities, intentional or unintentional. Common issues are improper codes, block billing, insufficient or vague descriptions, and math errors when indicating cost shares (sometimes in the firm’s favor). Undetected discrepancies can lead to a variety of issues such as over- or under-payment by the corporation, extra manual administration for both the law firm and corporation, lost attorney time trying to rectify errors, and damaged relationships between the firm and its client.

While e-billing tools have helped with identifying some invoice issues, most of the rules employed in these systems are both too crude to recognize nuances and too quick to throw alerts. Often, in-house attorneys’ time is wasted reviewing minor alerts from systems that trigger a significant number of false positive alerts while simultaneously failing to catch bigger issues such as overbilling (false negatives). 

The barrage of alerts generated by e-billing tools can quickly become burdensome for invoice reviewers and most switch back to manual review. The problem is, however, that reviewing an invoice often requires deep knowledge of context, the law involved, the history of the case, and familiarity with key parties to verify that the tasks performed and billed were necessary and appropriate. This means that final approvers may need to be high-rate attorneys or management in charge of the case. The effort required combined with the high volume of invoices results in either lost productivity due to the significant time and effort spent on manual review, or overpayment due to rushed invoice approval without proper review.

In this paper, we employ anomaly detection techniques tuned for law firm invoices, mitigating some of the challenges mentioned to reduce the corporate legal department effort needed for review and approval. We describe our use of SME to generate synthetic anomalies which are representative of an anomaly type with high importance of detection (anomalies in the legal case lifecycle). Once a dataset containing representative synthetic anomalies is generated, we apply several well-known and readily available model architectures to this task. Methods for evaluating the performance of these models are detailed and utilized for reporting results. We then present experimental results using real-world data from invoice line-items created by law firm billing customers for their services. While this work is presented through its application to the legal industry, it may be applied more generally to other similar types of data and in use cases where data patterns of the anomalies are well-known.

\section{Background and Related Work}

\subsection{Global and Local Anomalies}

The significant depth and breadth of anomaly detection research offers many anomaly types. With regard to detecting anomalies in line-items - which often contain significant amounts of categorical data - it is useful to consider two categories of anomalies: global and local \cite{Schreyer2017}. \textit{Global anomalies} exhibit unusual or rare individual attribute values (across the set of available features). These types of anomalies typically relate to rare attributes \cite{Das2007} and, when flagged, are often the source of significant false positive alerts \cite{Schreyer2017}. \textit{Local anomalies} exhibit unusual or rare combinations of attribute values while the individual attribute values themselves occur more frequently. These types of anomalies - while more difficult to detect - can be tied to behaviors typical of those conducting fraud \cite{Schreyer2017, Sparrow2019}. We use these characterizations throughout this work. 

% Utility across a wide range of applications, anomaly types, and data types has resulted in a variety of anomaly detection approaches \cite{Chandola2009, Goldstein2016}. Historically, FDS systems have been limited in their effectiveness due to the heavy utilization of preset rules that are defined through consultation with subject matter experts \cite{Li2008}. Due to the simplicity of these rule-based approaches, many are still utilized today. More recently however, more complex methodologies which are both unsupervised (without labeled training data) and supervised (making use of labeled training data) have been explored and shown to be effective in a variety of use cases \cite{Chandola2009}. These methodologies - when applied individually - cannot mitigate the effects of all types of anomalous activity within a system, and multiple methods must be employed by organizations to capture a broad range of fraudulent behavior. 

\subsection{Supervised and Unsupervised Modeling}

Supervised classification models have been used in health care fraud detection \cite{Travaille2011} and other industries such as computer security, telecommunications, etc. \cite{Schultz2020, Schreyer2017} due to their ability to detect specific fraud schemes. A major advantage of supervised learning is that the labels used for model training can be easily interpreted by humans for discriminative pattern analysis. 

Supervised learning models, however, can’t always be applied. First, it can be prohibitively expensive or difficult for many organizations to obtained labeled training data \cite{Abdallah2016, deRoux2018}. Second, even labeled data may contain ambiguities which stem from loose labeling guidelines or varying interpretations between data labelers. This results in less robust model decision boundaries. Additionally, new fraud schemes, differing in distribution from normal data and considered anomalies, are not immediately detectable due to the lag between discovering and subsequently labeling data as representing fraud \cite{Travaille2011}. 

For this reason, research attention has also focused on unsupervised approaches for anomaly detection \cite{Abdallah2016, Das2007, deRoux2018, Chen2016}. Unsupervised approaches overcome the labeling challenge, however, they have challenges related to the interpretability of results and ease of application. Unsupervised approach results will always require subject matter expert interpretation, negating some of their benefits. Additionally, many different types of anomalies may exist within a dataset, which makes interpretation more onerous. The need to interpret the results of unsupervised approaches provides the potential for mischaracterizations, such as considering many anomalous clusters to be a single anomalous group. This contrasts with supervised approaches, where models are typically trained to identify a single type of anomalous behavior.

\subsection{Electronic Invoicing in the Legal Industry}

Instead of mailing paper invoices, most law departments now require law firms to submit billing information in electronic format using e-billing systems and according to corporate billing guidelines.

In order to standardize the categorization of legal work and expenses and facilitate billing analysis, the Uniform Task-Based Management System (UTBMS) codes \cite{Aba} were developed in the mid1990s \cite{UTBMS} through a collaborative effort by the American Bar Association Section of Litigation and the American Corporate Counsel Association, among others. UTBMS is a series of codes used to classify legal services performed by a legal vendor in an electronic invoice submission. Attorneys record their time and classify tasks and expenses using the appropriate code from the UTBMS code set. UTBMS codes can be useful for both law firm and legal department data analytics purposes, for example when reviewing how much time has been spent on each type of activity (e.g. the percent of time spent on L200 Pre-Trail Pleadings and Motions vs. L400 Trail Preparation and Trail).  
In more advanced analyses, UTBMS codes can reveal the ordering of the activities throughout a legal case, such as showing the typical distribution of work (recorded by time) given a legal case’s attributes. An individual line-item can be assessed as to its suitability in the distribution of work (which defines the typical legal case lifecycle).

\begin{figure}
\vspace*{-0.22in}
\includegraphics[width=\textwidth]{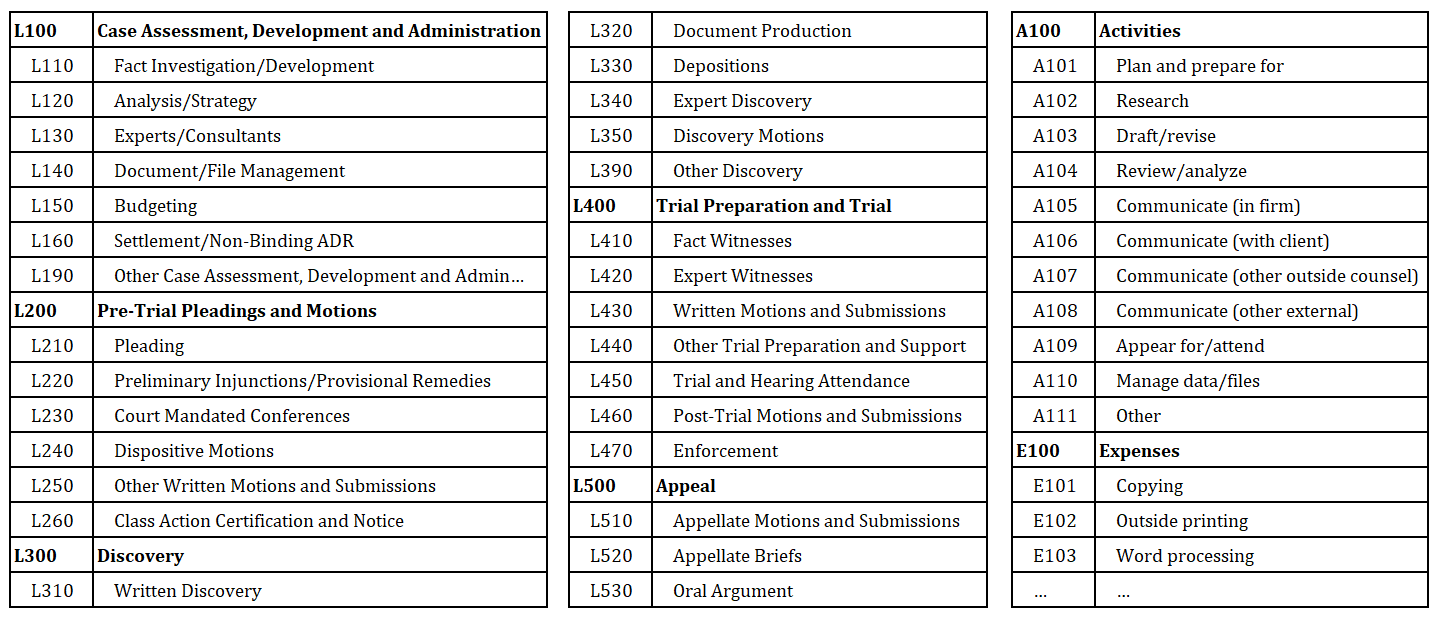}
\vspace*{-0.35in}
\caption{The above table shows available UTBMS codes for Litigation cases. The codes themselves are broken into phases, e.g. L100, L200, etc.} \label{utbms-codes}
\end{figure}

UTBMS codes offer many more benefits, however, there are some challenges. First, the use of UTBMS codes is inconsistent and largely dependent on each law firm’s individual timekeepers (e.g. attorneys, paralegals). Due to the sheer number of codes for different tasks in different phases on different case types, it is simply too difficult to memorize and utilize them effectively. Additionally, timekeepers may unintentionally assign incorrect codes because they are either too busy to search for the correct code or are indifferent to how these codes may benefit them or their clients. And in certain cases, some may intentionally bill fees or expenses with alternate codes to bypass e-billing alerts. When corporate legal departments do not recognize or acknowledge the negative impact of incorrect billing practices by their law firms, it may encourage timekeepers to continue ignoring or not following billing guidelines. As a result, on the corporate side,  billing and vendor management departments are left with muddled data containing many incorrect UTBMS codes, making accurate analytics and reporting a significantly more difficult task and negating many of the afore-mentioned UTBMS system benefits.

\section{Methods}

The following methods form the core components of a supervised anomaly detection approach that isolates preferred modeling data through dimensionality reduction and clustering, generates synthetic (anomalous) data based on information provided by SMEs, and utilizes well-known and easily accessible models for anomaly detection according to the characteristics of legal case lifecycles.

\subsection{Selecting Modeling Data}

A large corporation will typically hire law firms to provide legal services in different practice areas for open cases. These firms will submit invoices each month with line-items for each service provided. We sampled invoices with these line-items for our experiments, specifically line-items from \textit{Litigation} type cases (the most popular UTBMS case type). We applied a series of methods to select data that is suitable for the modeling process, i.e. data that would include the types of anomalies we were seeking to detect.

% \begin{table}[]
% \vspace*{-0.1in}
% \centering
% \begin{tabular}{@{}|lllllllp{3.6cm}|@{}}
% \toprule
% \textbf{Date} & \textbf{Type} & \textbf{Timekeeper} & \textbf{Qty.} & \textbf{Rate} & \textbf{Code} & \textbf{Activity} & \textbf{Description}                                                                                                   \\ \midrule
% 1/4/20      & Task          & John Doe            & 0.75              & \$350         & L100          & A108              & \begin{tabular}[c]{@{}l@{}}Meeting with all parties \\ regarding final version \\ for signoff approval\end{tabular} \\ \midrule
% 1/4/20      & Expense       & John Doe            & 1                 & \$201.50      & E110          & -                 & \begin{tabular}[c]{@{}l@{}}John Doe; Mediation \\ in Cleveland, Ohio.\end{tabular}                         \\ \bottomrule
% \end{tabular}
% \newline
%  \caption{An example of line-items contained within legal invoices. An invoice contains several line-items which may relate to activities or expenses. Hourly-rate invoices typically contain many line-items, while fixed-fee contracts often result in invoices with a single line-item to capture all the work in the contract.}\label{category_codes}
% \end{table}

% TODO: insert figure! 

Anomalies of interest to our study include line-items which, given the set of categorical attributes for the line-item, fall outside of the typical distribution of services billed in the legal case lifecycle. Cases that do not utilize codes spread across typical case phases may be ill-defined in the context of lifecycle anomaly detection as there is limited billing information. Such cases are often small in scope or settled early in the case lifecycle and do not cycle through later stages. These cases cannot contain an anomalous invoice line-item with respect to the case lifecycle, thus we removed these observations from the training set.

We then applied a two-step approach utilizing dimensionality reduction and unsupervised clustering to isolate Litigation type legal cases which are suitable for the modeling process. More specifically, we used features (generated from UTBMS codes) that indicate the distribution of billing charges across case phases. \textit{Phases} are higher-level categories defined in UTBMS codes (e.g. L100, L200, etc.). The following steps describe how the percentage of each phase charged per case is captured and processed. 

We used Singular Value Decomposition (SVD) \cite{Hestenes1958} and T-Distributed Stochastic Neighbor Embedding (T-SNE) \cite{Hinton2002} to reduce data dimensionality and visualize the charging patterns across cases. Subsequently we used Density-based Spatial Clustering of Applications with Noise (DBSCAN) \cite{Ester96} to group line-items according to related groups of charging patterns. Groups (clusters) containing Litigation type legal cases that exhibited well-distributed charging behaviors were retained for subsequent use in the experimental process, with “well-distributed” groups identified through a manual examination of the charging patterns (e.g. 60\% or more of charge codes utilized). Cases contained within groups of suitable data were utilized in the remainder of the experiment. All other cases were removed from the analysis. 

When applying SVD, we selected the number of principal components to use based on value, resulting in principal components that captured at least 95\% of the variation in the data. When applying T-SNE, a sufficiently low number of components were specified to simplify the clustering task. This approach combines the effectiveness of SVD with the interpretability benefits provided by T-SNE. By utilizing DBSCAN, we based the clustering results on the minimum distance between observations to be considered a “new” cluster, without the need to specify the number of clusters, and free from any assumptions about cluster distributions used by common approaches such as K-Means  \cite{Steinhaus1957, MacQueen1967, Lloyd1982} or Spectral Clustering \cite{Cheeger1969, Donath1972, Fiedler1973}.

\subsection{Synthetic Anomaly Generation}

Labeled training data was not available for our use case. To overcome this challenge, we utilized available SME to generate synthetic lifecycle anomalies in the line-item data which are representative of the types of anomalies we were seeking to detect. \textit{Lifecycle anomalies} are defined as anomalies which, with the exception of the number of days since the start of the legal case, are otherwise normal. That is, the line-item is anomalous with respect to the time in which it is present in the life cycle of the legal case and not according to its attribute values. Since we suspected that a very small number of these anomalies exist in the dataset, we employed an approach which generates synthetic anomalies to provide a basis for training supervised models specific to this task, similar to work performed by Schreyer et al.  \cite{Schreyer2017}.

First, we used the previously mentioned characterizations of global and local anomalies to identify global anomalies from the modeling dataset. Any combination of specified categorical feature (variable) values that occurs less than a specified number of times in the dataset can be considered rare occurrences, or global anomalies. These line-items were ignored later in the process. Lifecycle anomalies - whose individual values are not rare alone, but which are rare in combination - are considered local anomalies, like our anomalies of interest.

Once the rare and unusual line-items were marked, we generated a specified number of anomalies by manipulating the feature indicating the number of days since the legal case was opened. Until the specified number of anomalies was generated, a randomly selected line-item was drawn from the modeling dataset and adjusted so that the only feature out of place (given the other features) was the number of days since the case opened (more specifically, minimum and mean values for the days since the case started is calculated). We then utilized those values to modify line-items whose combination of feature values never occurs too early or too late in the lifecycle. The number of days since the case started was then adjusted for the selected line-item, such that it occurs far earlier in the lifecycle than is typical for the line-item’s combination of feature values. This results in a set of local anomalies, together with normal data, to utilize for model training and evaluation. These anomalies are considered local as the feature values themselves are common, but the number of days since the legal case started in combination with the feature values is considered out of place.

\subsection{Model Training and Testing}

\begin{wrapfigure}{c}{5.5cm}
\vspace*{-0.15in}
\centering
\includegraphics[width=5.5cm]{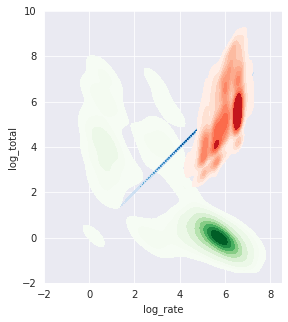}
\caption{The results of applying Gaussian Mixture Models (GMMs) for clustering of charging data.} \label{rate_total_clusters}
\vspace*{-0.1in}
\end{wrapfigure}

We utilized Random Forest \cite{Breiman2001}, Gradient Boosted Tree \cite{Breiman1997}, and Support Vector Machine (SVM) \cite{Cortes1995} model architectures in our experiments as these architectures are widely available, well-understood, and relatively performant across a broad variety of use cases. Additionally, we sought to apply a set of varying model architectures to provide several points of comparison.

Decision trees are widely employed given their simplicity, interpretability, and intelligibility. These models - using tree structures - represent class labels as leaves and conjunctions (combinations) of features that lead to those labels as branches. Decision tree modeling can however tend to overfit the data \cite{Hastie2008}. Random Forest Classifiers aim to overcome this tendency and reduce the variance in the predictions by constructing many decision trees (a ”forest”) and taking the mode (classification) or mean (regression) of the individual trees. These trees are fully grown and independently trained. Additionally, Random Forests provide a robust means to gauge the importance of features through approaches like Shapley Additive Explanations (SHAP) \cite{Lundberg2017} - a useful feedback mechanism which we used to guide feature selection \cite{Breiman2001}. Yet, Random Forests may not be as performant as Gradient Boosted Trees in some use cases, which aim to improve performance over decision trees in another way. 

Gradient Boosted Trees ensemble a set of weak learners, or shallow trees, in a bid to improve performance through a reduction of bias. Boosted trees train each weak learner sequentially, with each learner aiming to improve the performance of the ensemble of trees. With boosting, the training of a learner in the ensemble is dependent on the learners trained prior (in contrast with Random Forests). Due to the differences in their approaches for reducing prediction error, we elected to use both architectures in the experiments. 

For feature selection, we used a combination of SME knowledge and feature (variable) importance measures  \cite{Lundberg2017} to inform our final feature specification. We used both case-level and line-item-level attributes (encoded as features), as well as features which are derived through an earlier exploratory data analysis process. More specifically, we generated a feature which utilizes charging information (the hourly rate and the total) to distinguish groups of line-items by how they are billed (e.g. hourly or as a single invoice for the entire case) through applying a Gaussian Mixture Model (GMM) \cite{Amendola2016}. Additionally, we utilized the generated feature indicating the number of days since the legal case start (mentioned previously) but applied a log transformation to the feature before use in the models.

% \subsection{Inference on Unseen Data}

%% TODO: lastly, want to test behavior on dataset without labels. Flips exercise. Validation of model output through SME. 

%% TODO: model ensemble (majority vote?)

\section{Experiments}

For many anomaly detection systems, the performance of the selected approaches is difficult to assess. System assessments may be performed manually or through direct application to the use case. Fortunately, synthetic data representing anomalies of interest was generated and included in our experiments. Performance metrics can be directly computed by assigning the labels provided through synthetic data generation.

As described earlier, we used Litigation type cases in our experimental process. We conducted a grid search to identify the best possible parameters for each model and subsequently trained them. To provide a means of assessing performance, we classified each observation between two classes: \textit{anomalous} and \textit{normal}. Performance metrics were then calculated based on the consistency between the labels and the model’s predictions.

\subsection{Setup}

For a set of data containing one or more anomalous invoice line-items, we evaluated the last 20\% of the data - deemed a validation set - that is unseen by trained models. A model is trained for each model architecture according to a set of parameters identified as part of a grid search and a set of predictions is generated against the validation set. The performance of the model is assessed according to the agreement between the labels and predicted values provided by each trained model.

For each model architecture, the predicted classes for each line-item in the validation set was evaluated against the labels provided by the synthetic data generation according to the following rules:

\begin{enumerate}
  \item A \textbf{true positive} is recorded if the true label \textit{y} is equal to the predicted value $\hat{y}$.
  \item A \textbf{false negative} is recorded if the true label \textit{y} is anomalous but the predicted value $\hat{y}$ is estimated to be normal.
  \item A \textbf{false positive} is recorded if the true label 
 \textit{y} is normal but the predicted value $\hat{y}$ is estimated to be anomalous.
\end{enumerate}

We utilized the Python programming language for our experiments due to the wide variety of open source libraries available dedicated to machine learning and its general popularity with data scientists \cite{JetBrains2019}. Scikit Learn \cite{scikit-learn} was utilized for dimensionality reduction, clustering, and model training. The libraries Matplotlib \cite{Hunter2007}, seaborn \cite{waskom2020seaborn}, and Plotly \cite{plotly} were utilized for visualization, with Pandas \cite{mckinney-proc-scipy-2010}, Numpy \cite{Harris2020array}, and Scipy \cite{2020SciPy} also being utilized in our experimental pipeline.

\subsection{Model Parameters and Evaluation}

1,657 legal cases were represented in the originating dataset, with 973 Litigation type cases. The number of principal components for use in SVD was specified as 5, capturing approximately 97.91\% of the total amount of variation in the data. When applying T-SNE, the number of components was specified as 2, with the perplexity, learning rate, number of iterations, and the embedding initialization specified as 25, 50, 10000, and Principal Components Analysis (PCA), respectively. When applying DBSCAN, the maximum distance between two samples was set to 4.5 and the minimum number of samples in a neighborhood for a point to be considered a core point was set to 10. 

A threshold must be set that identifies global anomalies in a dataset from all others. When generating the synthetic anomalies used for model training, we utilized a threshold of 25, with any combination of feature values occurring less than the threshold considered to be global anomalies. We utilized a total of 5 features for this task related to the case category and line-item UTBMS code, type, and other information. In order to provide a suitable dataset for model training, we specified that 5\% of the newly-generated dataset contain local lifecycle anomalies.

For each model architecture, we utilized a common set of both categorical and numerical features that provide both case-level and line-item attributes. Some features included in the modeling process were the line-item-level UTBMS code, the activity, and the days since the case was opened. Case-level features included the case category and the number of unique UTBMS codes utilized. We evaluated the models according to the following performance metrics:
 
\begin{enumerate}
  \item \textbf{Precision:} The number of true positives \textit{tp} over the number of false positives \textit{fp} and true positives \textit{tp}, \( \frac{\textit{tp}}{\textit{tp} + \textit{fp}} \). 
  \item \textbf{Recall:} The number of true positives \textit{tp} over the number of false negatives \textit{fn} and true positives \textit{tp}, \( \frac{\textit{tp}}{\textit{fn} + \textit{fp}} \). 
  \item \textbf{F1 Score:} The harmonic mean of the precision and recall, defined as \newline 2 * \( \frac{precision * recall}{precision + recall} \) .
  \item \textbf{Accuracy:} The number of true positives \textit{tp} over the total number of samples in the validation set \textit{n}.
  \item \textbf{Coverage:} The number of samples which fall above a specified prediction confidence threshold (set to 90\%) over the total number of samples in the validation set \textit{n}. 
\end{enumerate}

The best-performing set of parameters for each model were identified and selected through a grid search to maximize the F1 score. The best set of parameters for each model architecture were recorded and applied to subsequent steps.

The modeling data is split into training, testing, and validation sets representing 60\%, 20\%, and 20\% of the data, respectively. More specifically, the training set represents the first 60\% of values in the data, with the validation set covering the last 20\%. Once the data was split into three distinct sets, we used time series cross-validation to train the model on the training data and assess its performance on the test data before generating our final set of predictions for evaluation on the validation set. We used the set of performance metrics defined above to evaluate and report the performance of each model.

%% todo: as applied to unseen data 

\subsection{Results and Discussion}

\begin{wrapfigure}{c}{6cm}
\vspace*{-0.4in}
\includegraphics[width=6cm]{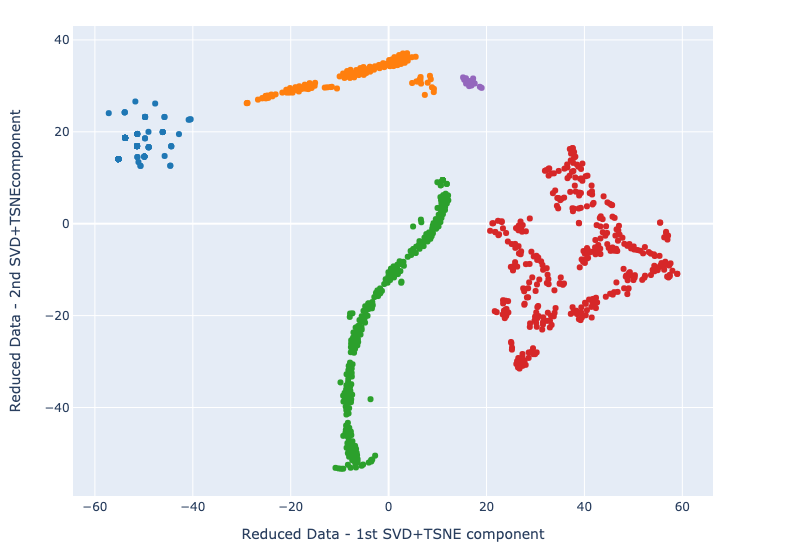}
\caption{Cluster 3, shown in red (farthest right), was identified as suitable for modeling.} \label{tsne}
\vspace*{-0.2in}
\end{wrapfigure}

\begin{wrapfigure}{c}{3.5cm}
\vspace*{-0.35in}
\centering
% \begin{figure}
\includegraphics[width=3.3cm]{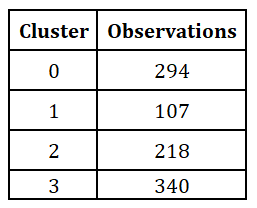}
\caption{The number of observations assigned to each cluster.} 
\label{clusters}
\vspace*{-0.25in}

% \end{figure}
\end{wrapfigure}

The data selection process returned a total of 340 legal cases for experimentation. With the parameter settings for data selection set as previously defined, a total of 5 clusters (groups) of legal cases were identified (see Table \ref{clusters} detailing the number of observations in each cluster and Figure \ref{tsne} for a graphical representation). Once the clusters of legal cases were identified based on their charging patterns, cluster 3 was selected as suitable for the modeling process based on a manual verification of the charging patterns for the cases in the cluster. This cluster provided a set of cases whose charging behavior is well distributed across the case lifecycle based on manual inspection of the line-items in the cluster.

\begin{figure}
\centering
\includegraphics[width=9.5cm]{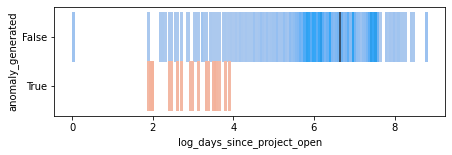}
\caption{The distribution of the log transform of the number of days since the case was opened across both non-anomalous (raw) data and anomalous (synthetic) data. The two distributions differ significantly.} \label{dist_log_days_since_open_flat_cat}
\end{figure}

The synthetic data generation process provided a dataset for model training and evaluation that included 1,967 anomalous line-items, with 77,305 line-items considered “normal”. Figure \ref{dist_log_days_since_open_flat_cat} illustrates the distributions of the log-transformed number of days since case start variable between the anomalous (class 1) and normal classes (class 0) in the dataset. As would be the case in a real-world scenario, the number of days since the case start values are significantly lower for line-items in the anomalous class versus those in the normal class. This follows the intuition that line-items which are anomalous with respect to the legal case lifecycle will exhibit regular (expected) feature values and be out of place only with respect to the chronology of when in the lifecycle they are charged. There is a clear and distinct separation between the two classes of observations (line-items) however, which can be easily detected by most machine learning algorithms. In future work, it may be suitable to explore a synthetic data generation strategy which results in a less distinct separation between the anomalous and normal classes. Models performing robustly under that scenario are likely to perform better overall. 

% \begin{wraptable}{r}{3.5cm}
% \vspace*{-0.2in}
% % \begin{table}[]
% \centering
% \begin{tabular}{|c|c|}
% \hline
% \textbf{Cluster} & \textbf{Obs} \\ \hline
% 0                & 294                   \\
% 1                & 107                   \\
% 2                & 218                   \\
% 3                & 340                   \\
% 4                & 14                    \\ \hline
% \end{tabular}
% \newline
% \caption{The number of observations (invoice line-items) assigned to each cluster.}\label{clusters}
% % \end{table}
% \vspace*{-0.2in}
% \end{wraptable}

% \begin{wrapfigure}{c}{12cm}
% \end{wrapfigure}

43,230 observations (line-items) were utilized in model training, with 14,410 and 14,411 observations belonging to the test and validation sets, respectively. As mentioned, time series cross-validation was used for model training and testing. This approach was selected due to its suitability in the context of how these models would be applied. 

Specifically, models would be periodically retrained on historical data and then employed for a specified amount of time to predict the classes of any new line-items in the system. Any data available up until the model is trained would be included as part of a pipeline, similar to the experimental process outlined in this paper. After that, features included in the model training process would be generated for any incoming line-items. These line-items would then be assessed for their anomalousness based on the earlier data.

\begin{table}
\vspace*{-0.1in}
\centering
\begin{tabular}{ |c||c|c|c|  }
 \hline
 \multicolumn{4}{|c|}{Validation Metrics on the Anomalous Class} \\
 \hline
Metric Name & Random Forest & Gradient Boosted Tree & Support Vector Machine\\
 \hline
Precision  & 91.16\% & 100\% & 74.35\% \\
Recall & 73.28\%  & 56.28\% & 98.30\% \\
F1-Score & 83.17\% & 72.12\% &  \textbf{84.67\%} \\
Accuracy & 97.82\% & 96.79\% & 97.38\% \\
Coverage & 97.29\% & 95.41\% & 93.76\% \\
 \hline
\end{tabular}
\newline
 \caption{A comparison between the performance of various model architectures on the experimental data. The best-performing model was used for each model architecture, identified through a grid search of possible parameters. It is important to recall that only 5\% of the data represents anomalies and that the F1-Score should be the most heavily considered metric.} \label{metrics}
\vspace*{-0.26in}
\end{table}

The performance metrics related to each model are shown in Table \ref{metrics}. Random Forest and SVM show comparable performance when assessed on the F1-Score yet differ in behavior. SVM provides more false positives but captures more  anomalies (high recall, low precision) while Random Forest returns a lower number of false positives at the expense of false negatives. SVM resulted in the highest overall F1-score at 84.67\% and also outperformed the other models in regards to recall (98.30\%). Gradient Boosted Tree showed the highest precision (100\%). While the tree-based methods were expected to perform relatively well based on their ability to capture non-linear relationships, their performance when compared to SVM is comparable and perhaps complementary if used in an ensemble. A model architecture could be selected by weighing the trade-off between higher false positive or false negative rates according to the specific anomaly detection problem. A high false positive rate could have significant adverse effects on end-user satisfaction \cite{LegalBillReview}. 

\begin{figure}
\vspace*{-0.2in}
\centering
\includegraphics[width=10.5cm]{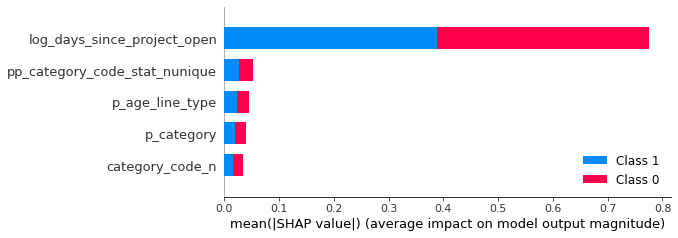}
\vspace*{-0.1in}
\caption{The Shapley Additive Explanations (SHAP) values from the Random Forest model for the top five most highly-weighted features. Note that the feature importances are weighed evenly across both the anomalous (class 1) and normal (class 0) classes.} \label{shap}
\end{figure}

We believe that there is value for future work in exploring additional feature engineering and model architectures which may yield improved performance beyond the results in this work, including exploring deep learning models for tabular data that utilize categorical feature embeddings. Feature selection was driven through a combination of SME and interpretation of Shapley Additive Explanations (SHAP) values \cite{Lundberg2017} provided as a result of modeling the Random Forest and Gradient Boosted Tree models. The feature importance values provided the feedback necessary to refine our earlier work and arrive at the feature set utilized in this paper. The top five feature importance values from the trained Random Forest model are shown in Figure  \ref{shap}. It is no surprise that the number of days since the case opened is weighed heavily by the model, as that is how this specific anomaly type (a line-item out of place in the legal case lifecycle) is defined.

%% TODO: results of feature importance plots and parameter selection
% \section{Deployment}

%% todo: workshop 

%% TODO: paragraph on how the methods in the paper are being utilized 

%% TODO: paragraphs on future work bases on the pilot deployment 

\section{Conclusion}

We provide an intelligent anomaly detection solution which works to address the challenges legal professionals face because of basic e-billing systems and misuse of UTBMS codes, raising costs and affecting overall efficiency. First, we highlight our approach to selecting data that is suitable for the experimental process and for lifecycle-type anomalies utilizing dimensionality reduction and clustering. Then, we provide a mechanism for applying supervised learning to anomaly detection problems in the legal industry when the behavior of potential anomalies is well-known. We highlight our approach towards generating a labeled dataset through manipulation of existing data in alignment with the concept of a local anomaly, providing a basis for training supervised models such as Random Forest, Gradient Boosted Tree, and SVM. The performance metrics for each of these model types against the synthetic data generation are provided and compared. An interpretation of the results and how they would inform application in the legal industry is then offered. By selecting a model architecture which minimizes false positives, this work could help further the flagging capabilities of e-billing software by minimizing erroneous alerts which reduce productivity. This work can improve the performance of automated alerts for e-billing systems in the legal industry.

\section{Acknowledgements}

This effort was supported by InfiniGlobe LLC located in Newport Beach, California, USA. The authors would like to thank Macon Mclean, Mike Russell (Expedia Group), Bobby Jahanbani (Exterro), Julie Richer (American Electric Power), Shahram Ghandeharizadeh (University of Southern California), Pasha Khosravi (University of California, Los Angeles), Monika Schreiner, and Dawn Kabiri for their feedback and support.

\bibliographystyle{splncs04}
\bibliography{references.bib}

\end{document}